\documentclass[letterpaper]{article} 
\usepackage{aaai25}  
\usepackage{times}  
\usepackage{helvet}  
\usepackage{courier}  
\usepackage[hyphens]{url}  
\usepackage{graphicx} 
\usepackage{subfigure}
\usepackage{natbib}  
\usepackage{caption} 
\usepackage{algorithm}
\usepackage{algorithmic}

\usepackage{amssymb}
\usepackage{amsmath}
\usepackage{txfonts}

\newcommand{\answerYes}[1][]{\textcolor{blue}{[Yes] #1}}
\newcommand{\answerNo}[1][]{\textcolor{orange}{[No] #1}}

\newcommand{\answerNA}[1][]{\textcolor{gray}{[NA] #1}}

\newcommand{\justificationTODO}[1][]{\textcolor{red}{\bf [TODO]}}

\usepackage{tikz}  
\usetikzlibrary{arrows,shapes,chains}  

\usepackage{soul, color, xcolor}

\usepackage{pifont}       
 
\usepackage{newfloat}
\usepackage{listings}

\DeclareCaptionStyle{ruled}{labelfont=normalfont,labelsep=colon,strut=off} 
\floatstyle{ruled}
\newfloat{listing}{tb}{lst}{}
\floatname{listing}{Listing}
\title{Constrained Behavior Cloning for Robotic Learning}

\author{
    Wensheng Liang\textsuperscript{\rm 1}\equalcontrib, 
    Jun Xie\textsuperscript{\rm 2}\equalcontrib, 
    Zhicheng Wang\textsuperscript{\rm 3}, 
    Jianwei Tan\textsuperscript{\rm 4}, 
    Xiaoguang Ma\textsuperscript{\rm 5}\thanks{Corresponding author.}
}
\affiliations{
    \textsuperscript{\rm 1}School of Mechanical Engineering and Automation, Northeastern University, Wenhua Road, Heping District, Shenyang, \\110819, Liaoning, China; 2200385@stu.neu.edu.cn\\
    \textsuperscript{\rm 2}Faculty of Robot Science and Engineering, Northeastern University, Chuangxin Road, Hunnan District, Shenyang, \\110169, Liaoning, China; 2402187@stu.neu.edu.cn\\
    \textsuperscript{\rm 3}Foshan Graduate School of Innovation, Northeastern University, Shunde District, Foshan, \\528311, Guangdong, China; 2390108@stu.neu.edu.cn\\
    \textsuperscript{\rm 4}Foshan Graduate School of Innovation, Northeastern University, Shunde District, Foshan, \\528311, Guangdong, China; 2410359@stu.neu.edu.cn\\
    \textsuperscript{\rm 5}Foshan Graduate School of Innovation, Northeastern University, Shunde District, Foshan, \\528311, Guangdong, China; maxg@neu.edu.cn\\
}

\usepackage{bibentry}

\begin{document}
\maketitle

\begin{figure*}[t]
    \centering
    \centerline{\includegraphics[width=\linewidth]{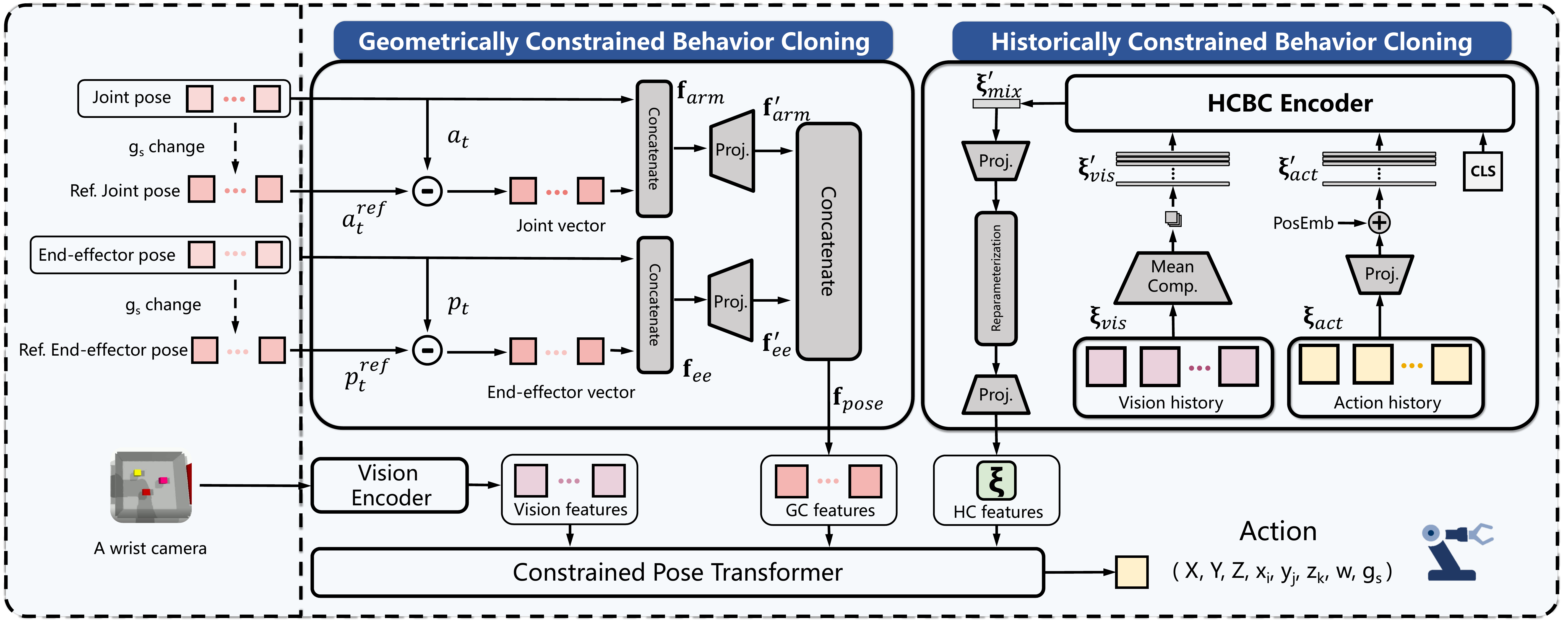}}
    \caption{The framework of geometrically and historically constrained behavior cloning (GHCBC).}
    \label{fig:1}
\end{figure*}

\begin{abstract}
Behavior cloning (BC) is a popular supervised imitation learning method in the societies of robotics, autonomous driving, etc., wherein complex skills can be learned by direct imitation from expert demonstrations.
Despite its rapid development, it is still affected by limited field of view where accumulation of sensors and joint noise bring compounding errors.
In this paper, we introduced geometrically and historically constrained behavior cloning (GHCBC) to dominantly consider high-level state information inspired by neuroscientists, wherein the geometrically constrained behavior cloning were used to geometrically constrain predicting poses, and the historically constrained behavior cloning were utilized to temporally constrain action sequences.
The synergy between these two types of constrains enhanced the BC performance in terms of robustness and stability.
Comprehensive experimental results showed that success rates were improved by 29.73\% in simulation and 39.4\% in real robot experiments in average, respectively, compared to state-of-the-art BC method, especially in long-term operational scenes, indicating great potential of using the GHCBC for robotic learning. 
\end{abstract}

\section{Introduction}
As the field of robotics advanced at an exponential pace, the concept of embodied intelligence had garnered significant interests and was recognized for its vast potential applications across various domains, including family service \cite{introduce1}, industrial automation \cite{introduce2}, education \cite{introduce3}, etc. However, the applications of embodied intelligence in practical robots required them to be able to handle multiple tasks with high precision within limited demonstrations. There were three main approaches for robots to imitate from demonstrations rapidly, i.e.,  inverse reinforcement learning (IRL) \cite{Eureka:journals/corr/abs-2310-12931}, imitation from human directly (IHD) \cite{MimicPlay:conf/corl/WangFSZ0XZA23}, and behavioral cloning (BC) \cite{RVT:conf/corl/GoyalXGBCF23}. The IRL learned a reward function from the demonstration data, so that the optimal strategy trained under this reward function could approach the demonstration data for practical applications. However, the reward function was usually difficult to design. The IHD method directly imitated actions through human demonstrations, and had the lowest data collection cost but with low accuracy. The BC method, on the other hand, could use demonstrations collected through teleoperating  robots to interact with environment for training. It aligned the environment and actions and allowed the robots to complete actions with higher precision.

Among these methods, BC's main advantages lay in its simplicity, speed, data efficiency, and stability, making it a preferred choice for quick deployment and efficient learning from expert demonstration \cite{introduce4}. However, collecting demonstration data in real world applications can be very difficult. For example, C-BeT \cite{FP2P:conf/iclr/CuiWSP23} required 4.5 hours of demonstration data to learn operational skills in specific scenarios. Additionally, the high cost of robots hindered the development of the BC as well. 
Recently, a low-cost open-source hardware system for bimanual teleoperation (ALOHA) \cite{ALOHA:conf/rss/ZhaoKLF23} was proposed to achieve dual arm manipulator teleoperation with an imitation learning algorithm, i.e., action chunking with transformers (ACT), wherein fine-grained biannual manipulation was realized by end-to-end BC from real demonstrations. Despite its surprising performance, the ACT observed environment with both internal and external cameras, making it unsuitable for universal manipulator with a single internal camera installed on the end-effector which was commonly used in practical applications. Furthermore, the ACT did not address the distribution shift issues which wa often caused by the uneven distribution of expert demonstration and policy execution in the BC \cite{introduce5, introduce7, introduce8}. This led to significant decrease in the model's reliability and stability.

Intuitively, integrating historical action information into robot BC could effectively reduce the negative effect of compound efforts and minimize the above-mentioned distribution shift issue \cite{introduce10}. In fact, human beings can learn more efficiently if they recall useful historical memories during imitation in most of their practical tasks \cite{introduce6}. By understanding past state information, the BC could effectively solve the limited distribution shift issues. Meanwhile, historical image information could also compensate for the limited field of view (FoV) of a single camera. Therefore, we could construct historical constrained behavioral cloning (HCBC) module, which integrated both vision history and action history to enhance temporal perception capabilities of the BC model. However, over-reliance on historical information could lead to copycat issue\cite{introduce9}, where robots woodenly replicated previous actions and ignored environmental changes. This made robot BC learning performance decrease with increasing integrated historical information and greatly reduced its flexibility and adaptability in dynamic environments. 

Practically, human brains usually pay more attention to relative pose information instead of details when learning from others imitatively \cite{introduce11}. For example, we learn how to use a keyboard by observing how others set up their relative pose information between hands and corresponding buttons on the keyboard, rather than memorizing the details of pitch, yaw, or roll of every single finger when they push the buttons. In fact, neuroscientists recently found that high-level perceptual information, such as relative pose information, was more dominant over low-level perceptual details, especially during associative memory recall of human brains \cite{introduce12}. This provided an inspiring way to solve the above-mentioned copycat issue. This paper designed a geometrically constrained behavior cloning (GCBC) module to integrate the state changes of robotic arms and end-effectors into the pose features to form geometrical constraints, corresponding to the high-level perceptual information in human brain learning. After introducing the HCBC and GCBC, the BC learning model not only effectively encoded historical information, but also paid attention to relatively important pose information and motion vectors, effectively mitigating the impact of the copycat issue. Experimental results showed that the Constrained Behavior Cloning method, which incorporated both HCBC and GCBC, achieved a 96\% success rate in controlled simulation experiments and a 92\% accuracy rate in real-world environments, significantly outperforming SOTA BC method.

Our main contributions of the paper were listed below:
\begin{itemize}
    \item Historical constrained behavioral cloning (HCBC) module was introduced to encode both vision and action history, enabling BC to learn temporal dependencies between current and past observations. 
    \item Geometrical constrained behavioral cloning (GCBC) module was developed to constrain predicted poses using pose vector traction methods, wherein high-level perceptual information such as relative pose information of joints and end-effectors was fully addressed to alleviate the copycat problem. 
    \item Comprehensive experiments on RLBench, simulations and real environments were conducted, wherein the GHCBC achieved 96\% and 92\% success rates, respectively, significantly surpassing SOTA BC method by 26\% in simulations and 30\% in real robot experiments, greatly enhancing its capabilities in practical few-shot imitation learning applications. 
\end{itemize}

\section{Related Work}
Behavioral cloning (BC) is the most widely used supervised imitation learning method, which can learn human-like policies for various tasks such as autonomous driving, robot manipulation, etc.
Despite its obvious advantages, it often encounters distribution shift issues due to differences between the training and testing data distributions leading to performance degradation \cite{wen2020fighting}.
High-end robots, equipped with multiple cameras and sensors (e.g., LiDAR, depth cameras), could have superior environmental perception adaptability, thereby mitigating the impact of distribution shifts \cite{bonci2021human}.
However, in practical applications, where robots typically have only a single front-end camera, the BC learning was a often unsatisfactory. 
The limited FoV provided by a single camera resulted in insufficient perceptual information and adaptability to environmental changes, especially when objects fell outside the FoV or lighting conditions varied abruptly \cite{davison2003real}.
As a result, robots often performed well during training but failed in real robot operations. This inconsistency undermined BC reliability and stability, necessitating extensive retraining and adjustments, which were time-consuming and costly \cite{schoppers1987universal}.

\subsection{Historically Constrained Behavior Cloning (HCBC)}
Traditional BC primarily relied on learning state-action pairs from expert demonstrations, often overlooking the significance of temporal sequence information during task execution. 
To enhance a model's awareness of task history, very recent research introduced state sequence encoders or historical trackers to better leverage past state information during decision-making.
For instance, \cite{mani2024diffclone} utilized historical trajectory data for decision modeling, demonstrating superior performance over traditional BC methods in diverse tasks.
\cite{kidera2024combined} combined constrained optimization with historical state aggregation to effectively reduce the negative impact of noise accumulation on model robustness.
Additionally, \cite{chi2023diffusion} employed historical data to predict future actions and used diffusion processes to minimize noise interference in decision-making.
Although the HCBC methods chould theoretically improve a model's temporal perception capabilities, an over-reliance on historical information often led to the copycat problem \cite{verbeek2013better}, where robots replicated observed actions after BC training, disregarding changes in the environmental context \cite{fighting-copycat:conf/nips/WenLDJ020}.
This limitation was particularly pronounced in dynamic or unstructured environments, significantly reducing the flexibility and adaptability of the BC learning.

\subsection{Geometrically Constrained Behavior Cloning (GCBC)}
According to neuroscientists, high-level state information should be dominantly considered for more efficient imitation learning.
In fact, human beings do focus more on relative pose information instead of specific details when learning tasks imitatively.
Inspired by this, we could mathematically design a GCBC module to realize integrating relative to geometrical information into the BC process to enhance accuracy and task execution capabilities.
It might need to leverage robot kinematics, visual geometrical features, and spatial constraints to improve BC learning performance in complex manipulation tasks.
Although the concept of the GCBC and its usage in robotics was introduced for the first time based on our best knowledge, there were some similar ideas appeared in this field.
For example, \cite{bicchi2000hands} utilized the spatial relationships between objects and their environments to assist models in making more precise decisions during complex manipulation tasks.
\cite{george2023one} combined action segmentation with a Transformer architecture to enhance their model's performance in multimodal tasks.
\cite{perez2021robot} integrated geometrical features from the environment to strengthen BC methods in robot navigation, thereby improving their capabilities in navigation and obstacle avoidance tasks.
Additionally, \cite{wang2023offline} introduced geometrical and diffusion processes to enhance strategy generation and multi-step prediction capabilities in complex tasks, particularly excelling in scenarios that required precise geometrical information.

\section{Methodology}
\subsection{Formulations of Behavior Cloning}
Behavior cloning (BC) aimed to learn a policy $\pi_\theta(a|s)$ that closely approximated the expert policy $\pi^*(a|s)$, where $s$ represented the state and $a$ denoted the action.
In this context, the expert policy $\pi^*(a|s)$ defined the probability distribution over actions $a \in \mathcal{A}$ given a state $s \in \mathcal{S}$. 
The training process involved constructing an expert demonstration dataset $\mathcal{D} = {(s_i, a_i)}{i=1}^N$ sampled from the training state distribution $\mathbb{P}_{\text{train}}(s)$.
The primary objective was to optimize the parameterized policy $\pi_\theta(a|s)$ to closely approximate the expert distribution, as defined by:

\begin{equation}\label{eq:1}
	\pi_\theta(a|s) \approx \pi^*(a|s).
\end{equation}

To achieve this, the BC typically minimized a log-likelihood loss function that quantified the divergence between the model's predictions and the expert's actions:

\begin{equation}\label{eq:2}
	\mathcal{L}_{\text{train}}(\theta) = - \mathbb{E}_{(s, a) \sim \mathbb{P}_{\text{train}}(s) \times \pi^*(a|s)} \left[\log \pi_\theta(a|s)\right].
\end{equation}

However, limited FoV inherent in the BC systems often caused a mismatch between the training state distribution $\mathbb{P}_{\text{train}}(s)$ and the test state distribution $\mathbb{P}_{\text{test}}(s)$, leading to performance degradation during testing.
For example, existing SOTA BC methods, including ACT \cite{ALOHA:conf/rss/ZhaoKLF23}, failed to adequately address the distributional shift issue caused by the uneven distribution of expert demonstrations and policy execution in the BC.
The resulting test loss was characterized by:

\begin{equation}\label{eq:3}
	\mathcal{L}_{\text{test}}(\theta) = - \mathbb{E}_{(s, a) \sim \mathbb{P}_{\text{test}}(s) \times \pi^*(a|s)} \left[\log \pi_\theta(a|s)\right].
\end{equation}

The presence of distributional shift, characterized by $\mathbb{P}_{\text{train}}(s) \neq \mathbb{P}_{\text{test}}(s)$, often led to suboptimal policy performance on test states that were underrepresented during training.

\begin{figure}[htbp]
    \centering
    \subfigure[Initial pose.]{
    \label{fig:left}
    \includegraphics[width=0.45\linewidth]{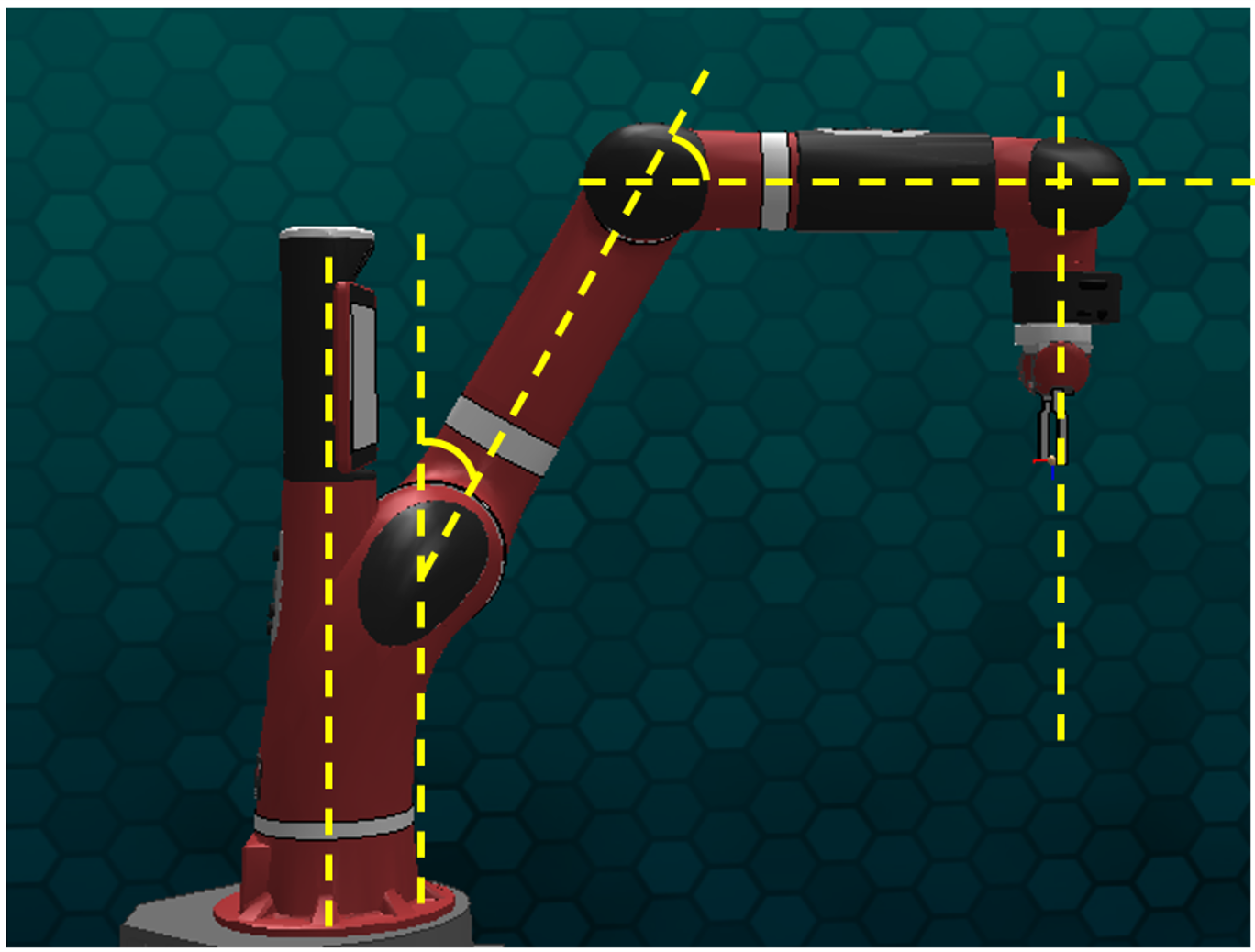}}
    \subfigure[Pose vector.]{
    \label{fig:right}
    \includegraphics[width=0.42\linewidth]{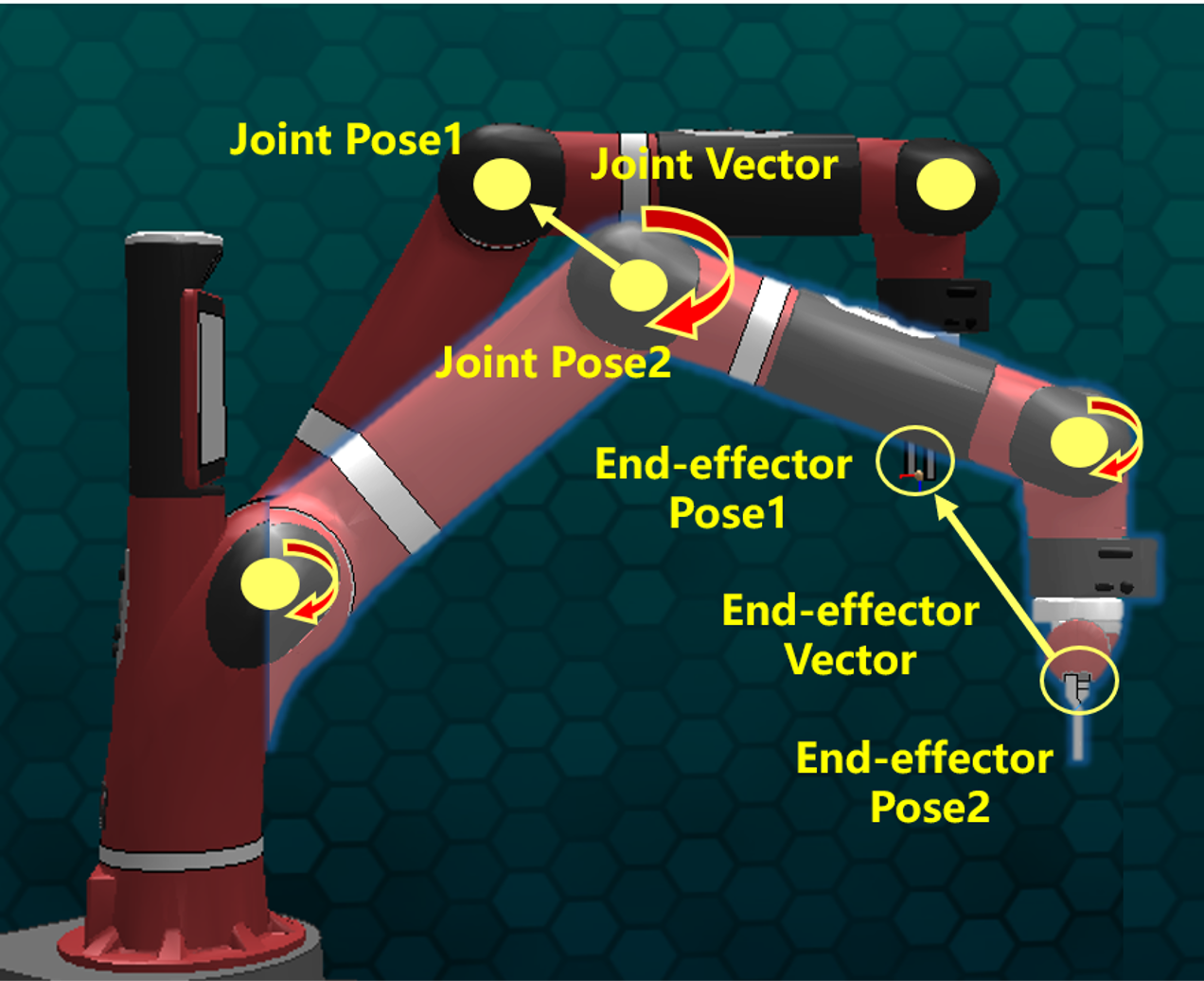}}
    \caption{Robotic arm pose and vector traction.}
\end{figure}

\subsection{Historically Constrained Behavior Cloning (HCBC)}
Historically constrained behavior cloning (HCBC) was introduced to address the challenge of performance degradation due to distributional shift by incorporating historical information $H_t = {(s_{t-1}, a_{t-1}), (s_{t-2}, a_{t-2}), \dots}$.
The HCBC policy $\pi_\theta(a|s_t, H_t)$ leveraged not only the current state $s_t$ but also past states and actions to construct a more robust BC learning model, to reduce compound errors.
However, excessive reliance on historical information $H_t$ could cause the policy to degenerate into $\pi_\theta(a|H_t)$, diminishing the importance of the current state $s_t$.
\begin{equation}\label{eq:5}
	\pi_\theta(a|s_t, H_t) \approx \pi_\theta(a|H_t).
\end{equation}

To mitigate this issue, the model incorporated the encoding and compression of action history information during the inference process, as shown in the upper right corner of  figure \ref{fig:1} . The following subsections outlined the implementation details:
\begin{itemize}
\item \textbf{Encoding of vision features}
Vision history data were encoded using a vision backbone network, processing a sequence of $k$ historical frames to generate feature representations $\xi_{\text{vis}} \in \mathbb{R}^{k \times 512 \times 20}$.
These features were then compressed using a mean compression layer, resulting in a simplified feature vector $\xi'_{\text{vis}} \in \mathbb{R}^{k \times 512}$ that was compatible with Transformer inputs.

\item \textbf{Encoding of action features}
The encoding of action history features produced historical motion posture information $\xi_{\text{act}} \in \mathbb{R}^{k \times 8}$.
A linear transformation was applied to map these features to a format consistent with the visual features, yielding $\xi'_{\text{act}} \in \mathbb{R}^{k \times 512}$.
Subsequently, positional encoding $\mathcal{P}$ was added to $\xi'_{\text{act}}$. Besides, there was an additional encoding tensor that participated in the model's training, updating its values during back-propagation. It transformed into higher-dimensional compressed encoding information, serving as a learned weight input to the model. The final output relied solely on the feature tensor encoded by the [CLS] token. Therefore, the features added into HCBC encoder could be obtained, i.e.,

\begin{equation}\label{eq:6}
	\xi_{\text{mix}} = \operatorname{Concat}([CLS], \mathcal{P}(\xi'_{\text{vis}}), \mathcal{P}(\xi'_{\text{act}})).
\end{equation}

\item \textbf{Integrated Encoding and Compression of Historical Information}
The concatenated sequence $\xi_{\text{mix}}$ was fed into a HCBC encoder, producing an output $\xi'_{\text{mix}} \in \mathbb{R}^{512}$.
To eliminate redundancy, this output was further compressed into a latent space representation $\xi_{\text{latent}} \in \mathbb{R}^{32}$ with $\mu_{\text{latent}} \in \mathbb{R}^{32}$ and $\sigma_{\text{latent}} \in \mathbb{R}^{32}$.
The encoding process was regularized using Kullback-Leibler (KL) divergence to ensure that the model output HC features retained only the most essential information:

\begin{equation}\label{eq:7}
	\mathcal{L}_{\text{KL}} = D{\text{KL}}\left(\mathcal{N}(\mu_{\text{latent}}, \sigma_{\text{latent}}) || \mathcal{N}(0, I)\right).
\end{equation}

After KL divergence, these features were passed through a linear projection layer, restoring the feature dimension to 512, and obtaining the final HC features $\xi$.
\end{itemize}

\subsection{Geometrically Constrained Behavior Cloning (GCBC)}
To overcome the limitations of Historically constrained behavior cloning (HCBC), we introduced Geometrically constrained behavior cloning (GCBC), which incorporated geometrical constraints $G(s_t)$ into the decision-making process, as shown in the upper left corner of figure \ref{fig:1}.
In this framework, the policy $\pi_\theta(a| G(s_t), H_t)$ was guided by the geometrical features of the current state $s_t$, rather than relying solely on historical patterns $H_t$.
This approach could enhanced the model’s adaptability to dynamic environments and ensure that current state information was adequately considered during action selection.

Practically, we implemented the geometrical constraints on robotic joint and its end-effector displacement, wherein pore vector traction (PVT) module was introduced to better consider high-level perceptual information of manipulators, as shown in figures \ref{fig:left} and \ref{fig:right}.
The GCBC consisted of two primary components:
\begin{itemize}
\item \textbf{Joint pose and joint pose vector}: The joint pose $a_t$ represented the rotational angles of each joint and gripper state, serving as the most direct observable pose information of the robotic arms. The joint pose vector $\Delta a_t$, defined as the difference between the current joint pose $a_t$ and the reference jointed pose $a^{ref}_t$, was given by:
\begin{equation}\label{eq:8}
	\Delta a_t = a_t - a^{ref}_t.
\end{equation}

Here, $a^{ref}_t$ was updated to the current $a_t$ whenever the gripper state changed ($g_s$ change), as shown in upper left corner in figure ~\ref{fig:1}.
This formulation could enhance the model's ability to perceive and predict the end-effector's state, thereby improving the stability of the BC learning.

\item \textbf{End-Effector Pose and End-Effector Vector}: The end-effector pose $p_t$ included three-dimensional position $(x_t, y_t, z_t)$, rotation quaternion $(x_i, y_j, z_k, w)$, and gripper state. The end-effector vector $\Delta p_t$ was defined as the displacement between the current end-effector pose $p_t$ and the reference pose $p^{ref}_t$:
 \begin{equation}\label{eq:9}
	\Delta p_t = p_t - p^{ref}_t.
\end{equation}

Similarly, $p^{ref}_t$ was updated to the current $p_t$ upon changes in the gripper state. This vector effectively represented the spatial displacement of the end-effector, providing a robust geometrical traction signal to constrain the BC learning.
\end{itemize}

\begin{figure}[t]
    \centering
    \centerline{\includegraphics[width=\linewidth]{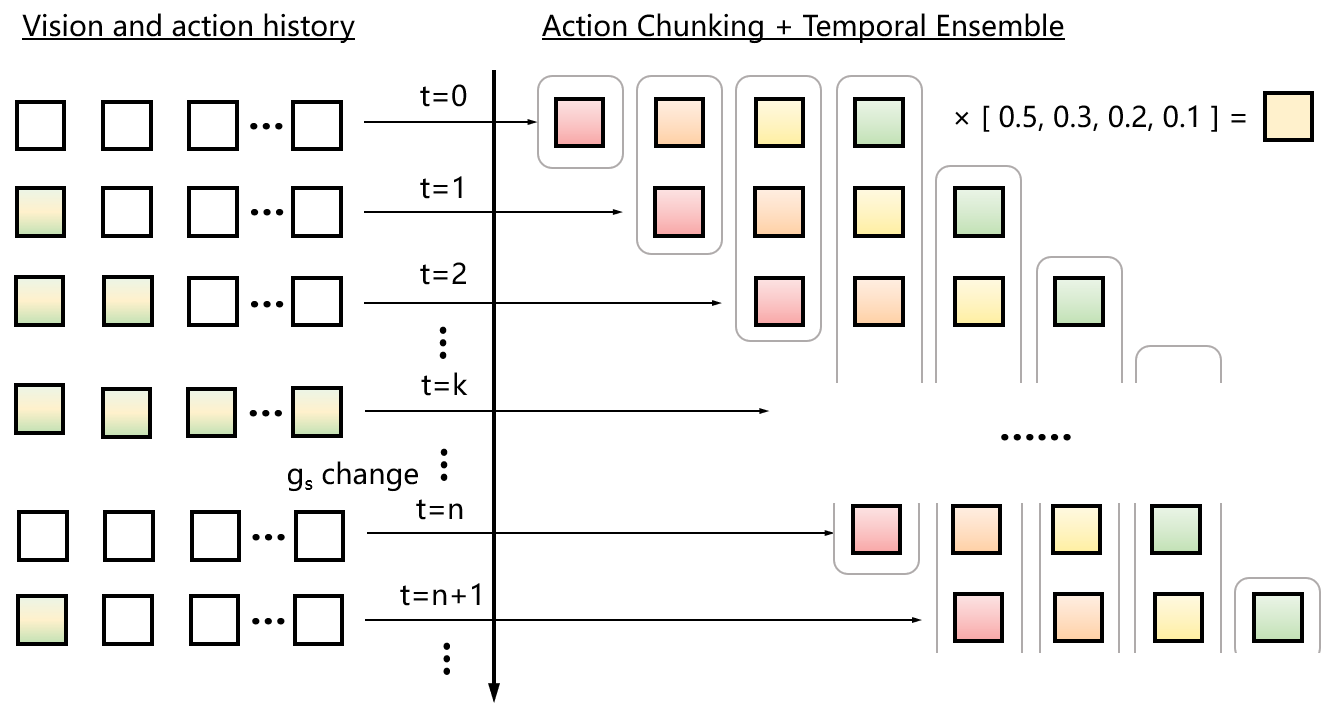}}
    \caption{History information with action chunking and temporal ensembling.}
    \label{fig:CPT}
\end{figure}

Additionally, when the gripper state changed, we cleared the vision and action histories to let the model focus on the most recent action history, thereby mitigating distributional shift, as shown in figure \ref{fig:CPT}. Specifically, when the gripper state changed, the history information was entirely replaced by padding information.To process the pose and vector features of the robotic arms and their end-effectors, we first concatenated the respective features into two one-dimensional tensors of length 15:
 \begin{equation}\label{eq:10}
	\mathbf{f}_{\text{arm}} = \text{Concat}(\Delta a_t, a_t), \quad \mathbf{f}_{\text{ee}} = \text{Concat}(\Delta p_t, p_t).
\end{equation}

These tensors were then individually passed through a linear scaling layer, transforming them from 15 dimensions to 512 dimensions,  thereby obtaining the tensors $\mathbf{f}'_{\text{arm}}$ and $\mathbf{f}'_{\text{ee}}$. Finally, the scaled tensors were concatenated to form the final GC feature representation:
 \begin{equation}\label{eq:11}
	\mathbf{f}_{\text{pose}} = \text{Concat}(\mathbf{f}_{\text{arm}}', \mathbf{f}_{\text{ee}}').
\end{equation}

\subsection{Detailed Architecture of the Gometrically and Historically Constrained Behavior Cloning (GHCBC) }
Figure \ref{fig:1} illustrated the detailed structural diagram of our proposed method, which included the GCBC and the HCBC modules. We first encoded the images captured by the wrist camera through a vision encoder to obtain vision features. The encoder model input was the 160×120 pixel image. The output feature map was a tensor with size of (1280, 4, 5), which was projected to (512, 4, 5) through a fully connected layer size of (1280, 512). A sine function position encoding tensor was then added to preserve spatial information. After position encoding, the feature map was flattened into (512, 20), and the dimensions were swapped to generate 20 tensors of length 512, which were used as the final vision features (20, 512).
Concurrently, manipulator sensors collected joint pose, end-effector pose, as well as stored reference joint pose and reference end-effector pose, which were then fed into the GCBC module.
By calculating and integrating joint vector and end-effector vector information, GC features were derived. HC features were inferred by inputting stored vision history and action history into the HCBC module. Subsequently, the vision features, GC features, and HC features were collectively sent to the Constrained Pose Transformer for action sequence prediction, and the action information required by the arm was obtained through action chunking and temporal ensembling operations, thus completing the inference task of the entire model. Finally, the vision features and action generated from this round of inference would be stored in the vision history and action history to use as history features in the next iteration. 

The post processing of GHCBC was shown in figure \ref{fig:CPT}, at the initial execution of the model, the vision history and action history did not reach the sequence length of k. In this case, we used padding to fill in the blank parts. The length of the history information was consistent with the set value of k. As the inference proceeds, we updated vision history and action history to vision first-in-first-out (FIFO) buffers and action FIFO buffers, respectively, to drop vision history and action history before k. 
This continues until the gripper state changed, at which point the history information is completely cleared and a new round of updates begins. The operations of action chunking and temporal ensembling were not affected by changes of the gripper and would continue to operate according to a fixed pattern.To better introduce our method, we also summarized the GHCBC training and GHCBC inference of our method in algorithms 1 and 2.

\section{Experiments and Discussions}
\subsection{Experiment setup}
\subsubsection{Device used for training}
We evaluated geometrically constrained behavior cloning (GCBC) and historically constrained behavior cloning (GCBC) module by conducting comprehensive experiments in both simulation and real-world scenarios. Both simulation platforms and the real world tests used Sawyer robots for evaluation. The GHCBC was trained and deployed with Intel Core i7-11700, 64GB of RAM, and NVIDIA GeForce RTX 3080ti 12G.

\begin{figure*}[t]
    \centering
    \centerline{\subfigure[]{\includegraphics[width=0.15\linewidth]{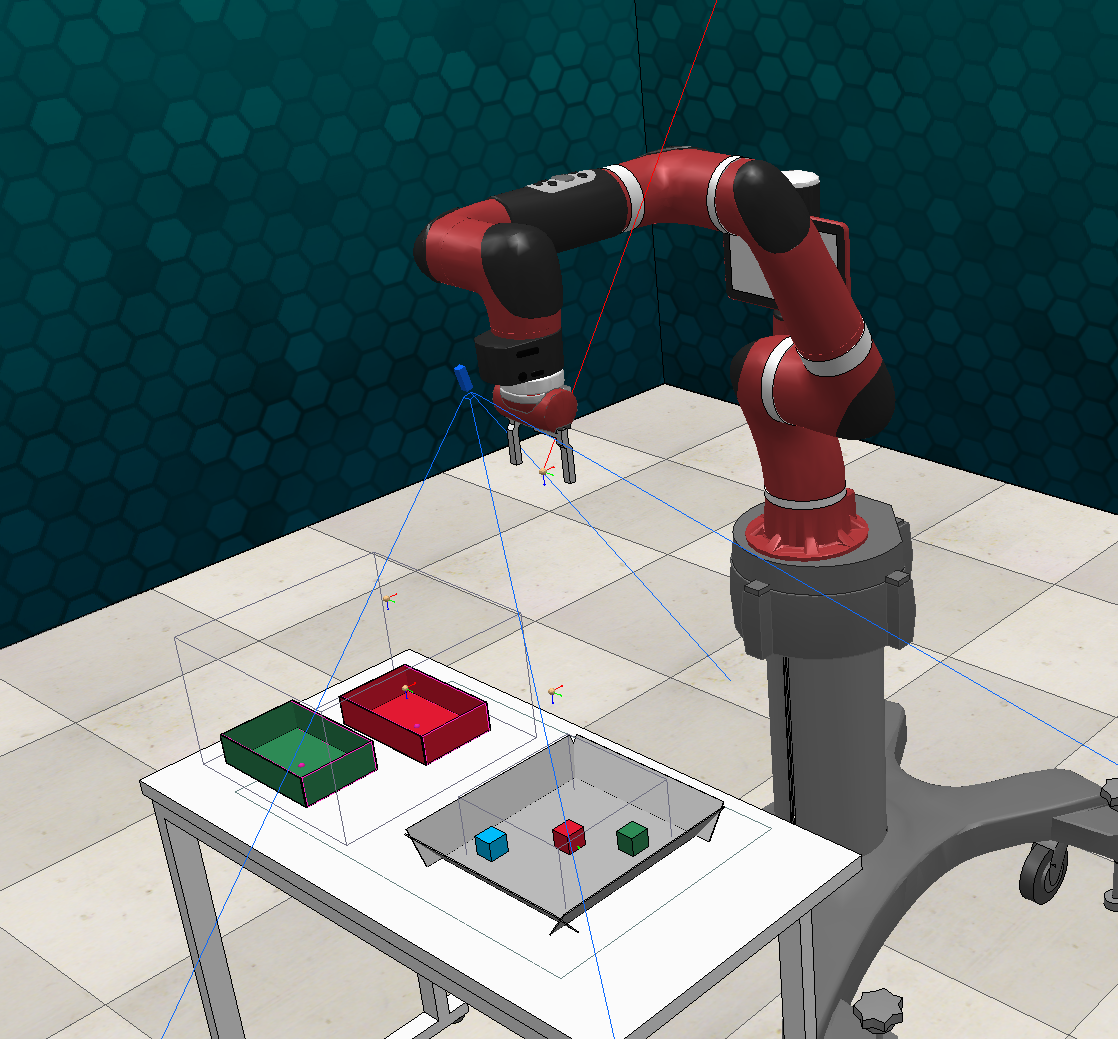}\label{task_show:a}}
    \subfigure[]{\includegraphics[width=0.14\linewidth]{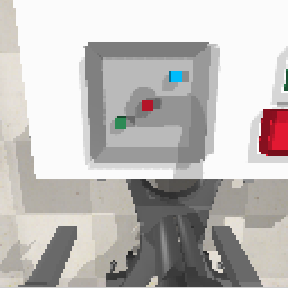}}
    \subfigure[]{\includegraphics[width=0.15\linewidth]{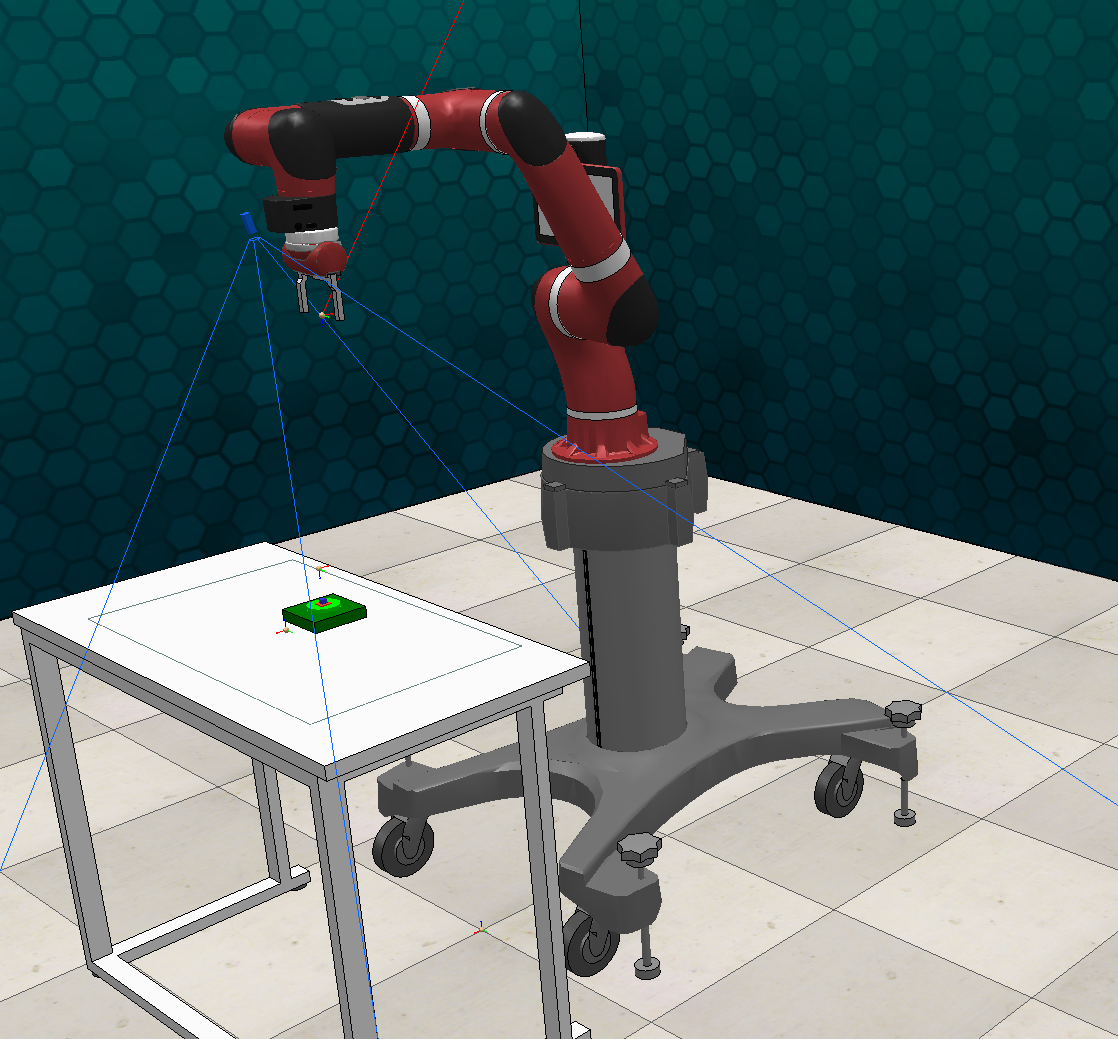}}
    \subfigure[]{\includegraphics[width=0.14\linewidth]{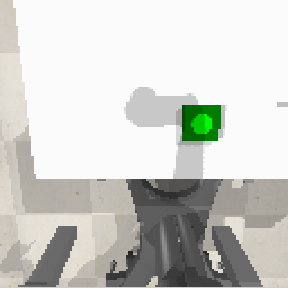}}
    \subfigure[]{\includegraphics[width=0.144\linewidth]{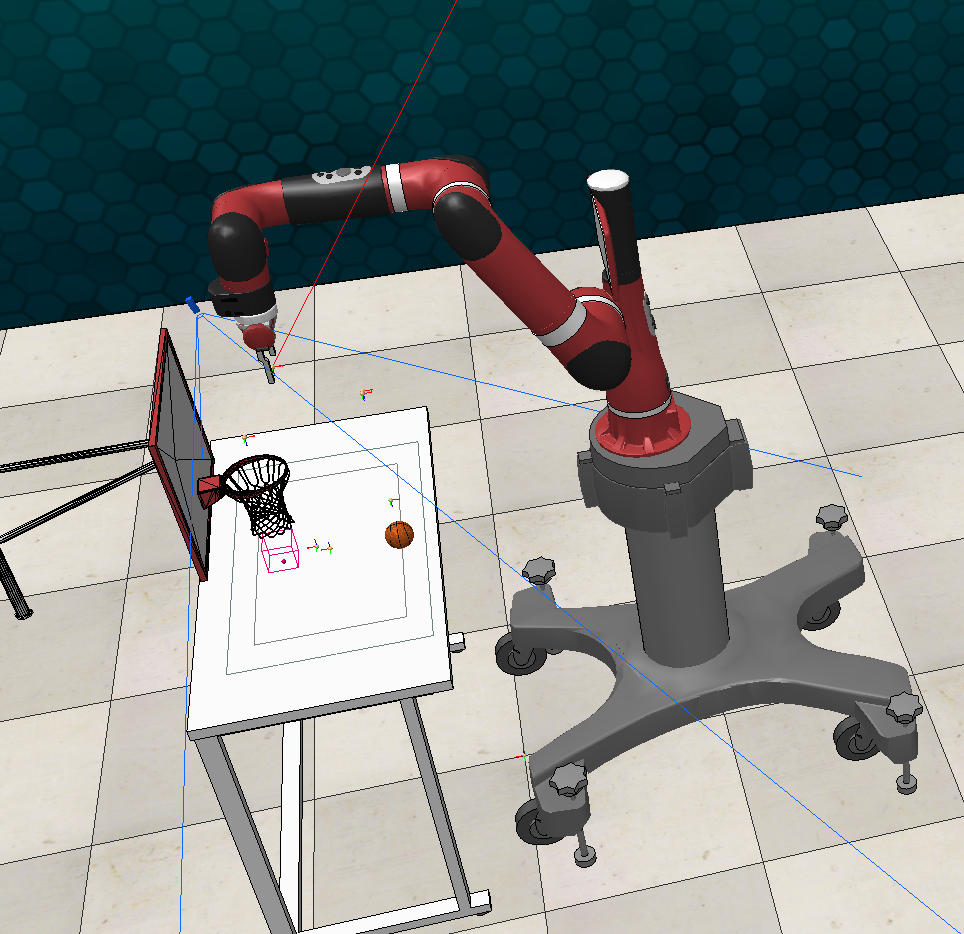}}
    \subfigure[]{\includegraphics[width=0.14\linewidth]{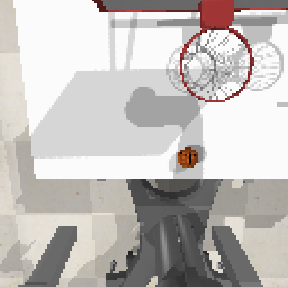}\label{task_show:i}}    \subfigure[]{\includegraphics[width=0.14\linewidth]{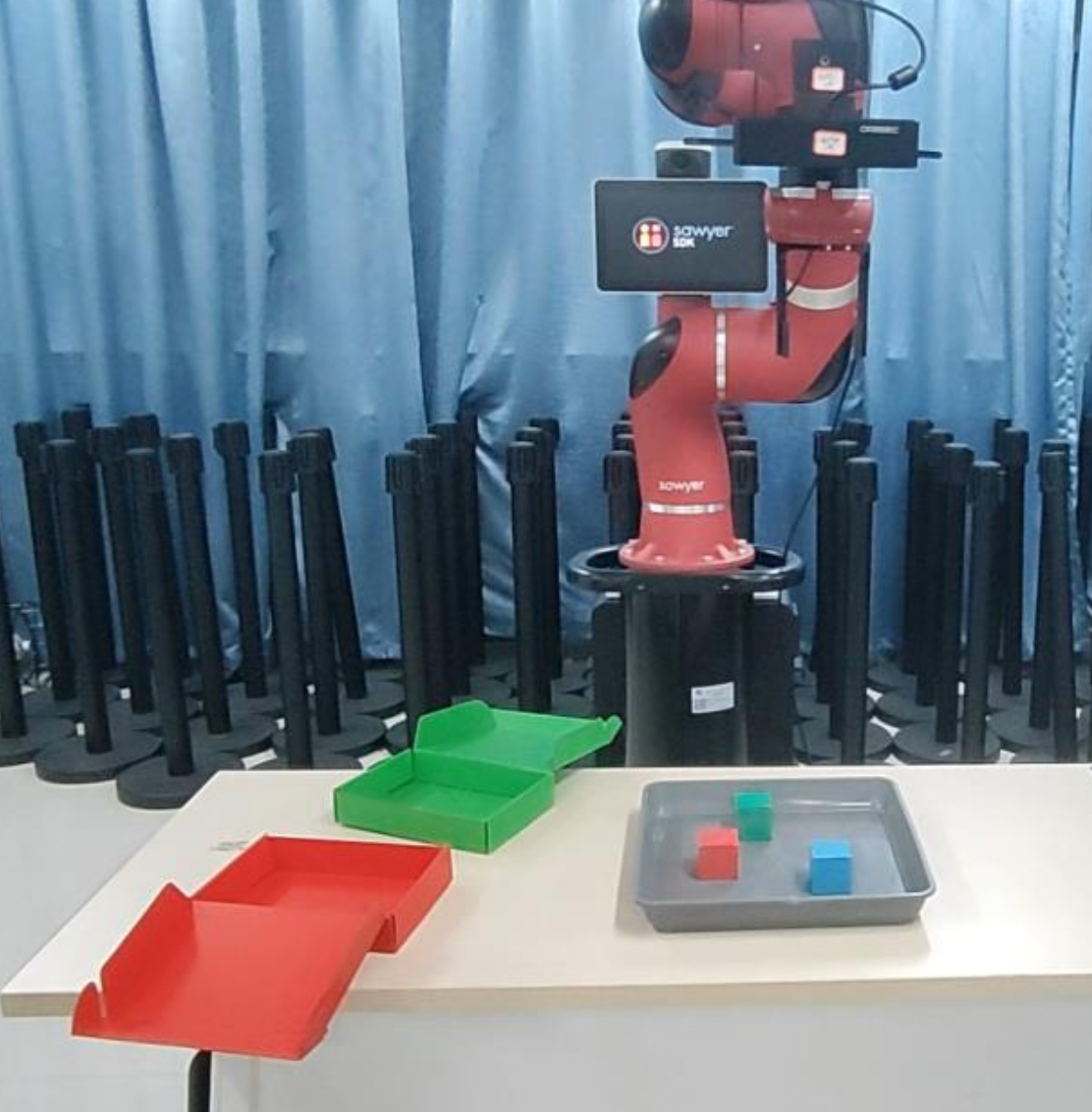}\label{task_show:real}}}
    \label{task_show}
    \caption{Simulation tasks setup. (a) Sorting task setup. (b) Image observation in initial position of the sorting task. (c) Button pushing task setup. (d) Image observation in initial position of the button pushing task. (e) Basketball shooting task setup. (f) Image observation in initial position of the basketball shooting task. (g) Sorting task setup in real world. }
\end{figure*}

\begin{figure}[h]
    \centering
    \subfigure[]{
    \label{hand_camera}
    \includegraphics[width=0.43\linewidth]{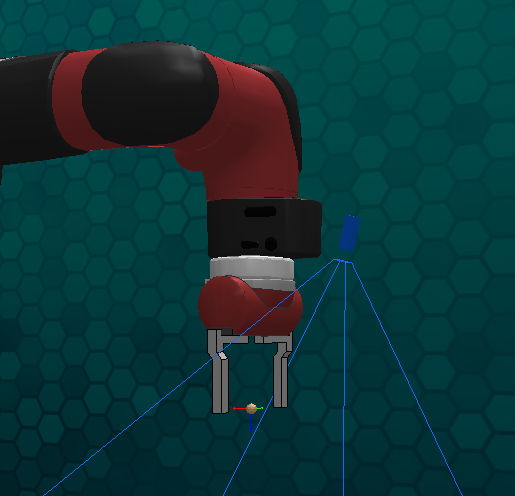}}
    \subfigure[]{
    \label{simset-multi-light}
    \includegraphics[width=0.45\linewidth]{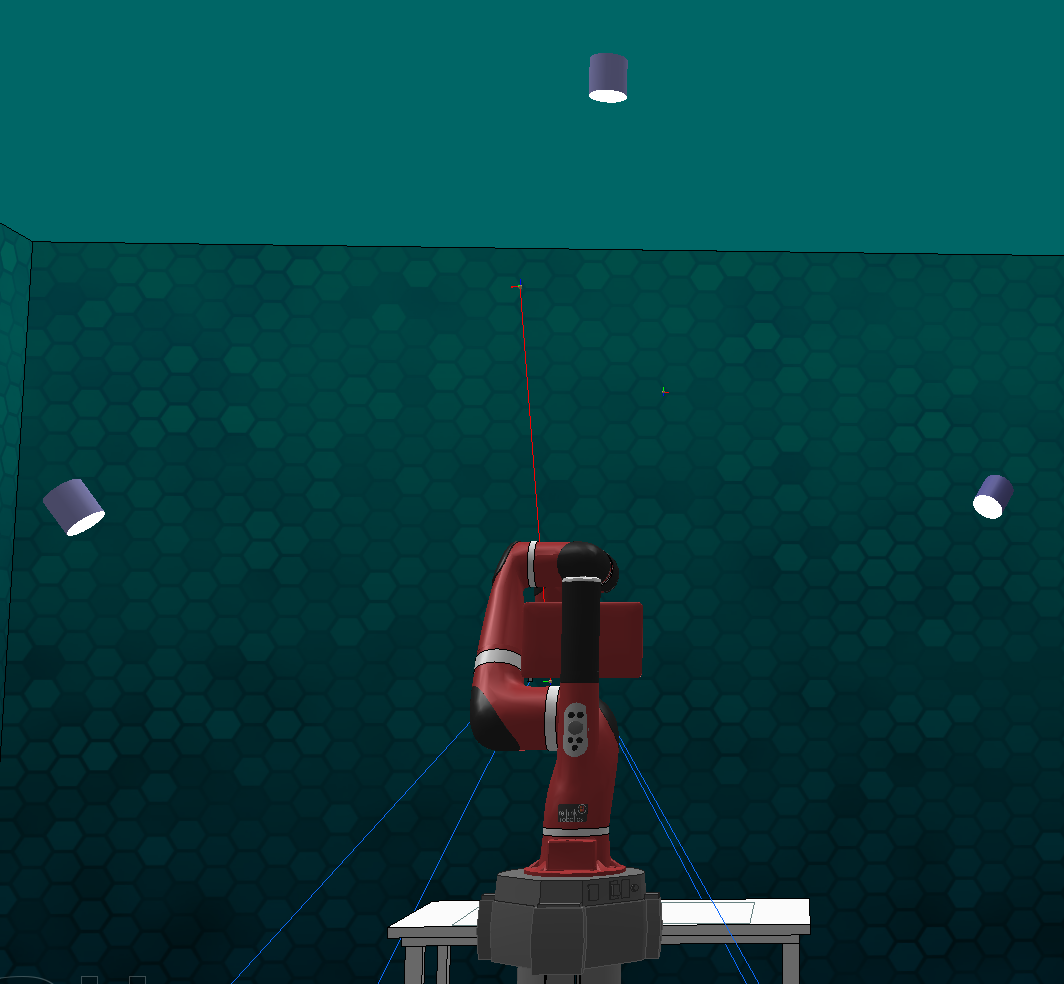}}
    \caption{Simulation environment set up. (a) Wrist camera installation pose related to end-effector. (b) Multi-light setup in simulation scene. }
\end{figure}
\begin{algorithm}[h]
    \caption{GHCBC Training}
    \label{alg:cbc_train}
        \begin{algorithmic}[1]
        \REQUIRE{Demo dataset $\mathcal{D}$, chunk size $k$, KL weight $\beta$}
        \STATE Let $a_t$, $j_t$, $e_t$, $i_t$ represent output pose, joint angles, end-effector pose, and image observation at timestep $t$. 
        \STATE Initialize HCBC Transformer Encoder $q_\phi(\xi | a_{t:t+k}, \Bar{o}_t)$.
        \STATE Initialize Constrained Pose Transformer $\pi_\theta(\hat{a}_{t:t+k} | o_t, z)$.
        \FOR{iteration $n=1, 2, \dots$}
            \STATE Sample $a_{t-k-1:t+k}$, $j_t$, $e_t$, $i_{t-k:t}$ from $\mathcal{D}$.
            \STATE Obtain $\hat{a}_{t:t+k}$ from Algorithm \ref{alg:cbc_inference}.
            \STATE $\mathcal{L}_{reconst} = \mathit{MSE}(\hat{a}_{t:t+k}, a_{t:t+k})$.
            \STATE $\mathcal{L}_{reg} = D_{\mathit{KL}}{\mathcal{N}(0,\mathit{I})}$.
            \STATE Update $\theta$, $\phi$ with ADAM and $\mathcal{L} = \mathcal{L}_{reconst} + \beta\mathcal{L}_{reg}$.
        \ENDFOR
        \RETURN{trained $\pi_\theta$ and $q_\phi$}

        \end{algorithmic}
\end{algorithm}
\begin{algorithm}[h]
\caption{GHCBC Inference}
\label{alg:cbc_inference}
\begin{algorithmic}[1]
    \REQUIRE{trained $\pi_\theta$ and $q_\phi$, episode length $T$, weight $m$}
    \STATE Initialize first-in-first-out buffers $\mathcal{A}[0:T]$ and $\mathcal{I}[0:T]$, wherein $\mathcal{A}[t]$ and $\mathcal{I}[0:T]$ store predicted actions and images \textit{for} timestep $t$, respectively.
    \STATE Sample $j_1$, $e_1$ from environment.
    \STATE Let $j^{ref}_r$, $e^{ref}_r$, $g^{ref}_s$ = $j_1$, $e_1$, 0.
    \FOR{timestep $t$ in $n=1, 2, \dots, T$}
        \STATE Sample $j_t$, $e_t$, $i_t$ from environment
        \STATE Sample $\xi$ from $q_\phi(\xi | \mathcal{A}[0:T], \mathcal{I}[0:T])$.
        \STATE Obtain GC features $\mathbf{f}_{\text{pose}}$ from GC transformation $p(j_t, e_t)$
        \STATE Predict $\hat{a}_{t:t+k}$ from $\pi_\theta(\hat{a}_{t:t+k} | \xi, \mathbf{f}_{\text{pose}}, i_t)$.
        \IF{\textit{training}}
            \STATE \textbf{return}$\ \hat{a}_{t:t+k}$
        \ENDIF
        \STATE Add $\hat{a}_{t:t+k}$ to buffers $\mathcal{A}[t:t+k]$ respectively.
        \STATE Add $i_{t}$ to buffers $\mathcal{B}[t:t+k]$ respectively. 
        \STATE Obtain current step actions $A_t = \mathcal{A}[t]$.
        \STATE \textbf{Apply} $a_t = \sum_{i} w_i A_t[i]/\sum_{i} w_i$, with $w_{i} = \exp(-m*i)$.
        \STATE Obtain gripper state $g_s$ from $a_t$. 
        \IF{round($g_s$) $\neq$ $g^{ref}_s$}
            \STATE clear $\mathcal{A}[0:T]$.
            \STATE clear $\mathcal{I}[0:T]$.
            \STATE Let $j^{ref}$, $e^{ref}$, $g^{ref}_s$ = $j_t$, $e_t$, round($g_s$).
        \ENDIF
    \ENDFOR
\end{algorithmic}
\end{algorithm}

\subsubsection{Dataset setup}
We used RLBench \cite{9001253}, a widely used large-scale benchmark for promoting research on vision guided robot manipulation, to generate expert demostrations of 3 tasks, i.e., a button pushing task, a basketball shooting task, a block sorting task, as shown in figure \ref{task_show:a} to \ref{task_show:i}, and a block sorting task in real world, as shown in figure \ref{task_show:real}. 
In the sorting task, the robot should pick up a specific block and place it into a specific box. The simulation sorting task mainly included a Sawyer robot fixed on the base, a long table, a tray, three different colored blocks, and two different colored boxes, wherein the positions and angles of the different colored blocks and boxes were randomly changed within a certain color set, as shown in figure \ref{task_show:a}. In the button pushing task, the robot should push a randomly generated button on the desk. In the basketball shooting task, the robot should pick up a randomly generated basketball on the desk and put it in the hoop. A 7-DoF Sawyer robot with a parallel gripper was utilized to complete the tasks. We extracted the model of the Sawyer robot from CoppeliaSim's robot structure library, and forward and inverse kinematics calculation configurations from PyRep \cite{james2019pyrepbringingvrepdeep}. We added a wrist RGB camera on its end-effector only to capture the robot's visual inputs to evaluate the performance of the HCBC and the GCBC, as shown in figure \ref{hand_camera}. 

\subsubsection{Scene deployment}
In order to simulate the multi-light situation indoors or outdoors, three ordinary linear light source scene lighting were set in the task scene, and the shadow effect under multi-light sources was simulated, as shown in figure \ref{simset-multi-light}.
The initial posture of the Sawyer robot joints was (0, -60, 0,+60, 0, 90, 0) to ensure that objects on the table could be seen.
The target object used in this experiment was a block with a side length of 40mm. Meanwhile, there were two interfering blocks in the sorting tray, wherein their positions, angles, and colors were also random chosen, to verify the model's ability. The colors of the block could be randomly selected from 7 actual colors available in the real experiment, in order to maintain consistency between the simulation environment and the real world environment as much as possible. Other parameters of the simulation were listed in table \ref{para-table}. 
\begin{table}[h]
    \setlength{\tabcolsep}{0pt}
    \centering
    \scalebox{0.9}{
        \begin{tabular}{cc}
            \hline
            Parameters & Values \\
            \hline
            Resolution of the wrist camera & 120 $\times$ 160 \\
            \begin{tabular}{c}Range of random center \\ positions of blocks\end{tabular} & 250mm $\times$ 250mm\\
            Minimum distance of blocks & 10mm\\
            Size of blocks & 40mm $\times$ 40mm $\times$ 40mm\\
            \begin{tabular}{c}Range of random center \\ positions of boxes\end{tabular} & 350 $\times$ 350\\
            Minimum distance of boxes & 10mm(no collision)\\
            Size of boxes & 50mm $\times$ 125mm $\times$ 200mm\\
            \hline
        \end{tabular}
    }
    \caption{Environment parameters.}
    \label{para-table}
\end{table}

\begin{table*}[h]
    \centering
    \setlength\tabcolsep{2pt}
    \begin{tabular}{ccccccccccccc}
    \hline
   Row ID & Pose input & Pose output & Policy trainer & Sim Succ. rate(\%) & Real Succ. rate(\%)\\
   \hline
   1(ACT)       & Joint & Joint & Style variable & 70  & 62\\
   2                 & Joint & End-effector& Style variable & 74  & 66\\
   3                 & Joint + End-effector & End-effector & Style variable & 60  & 56\\
   4                 & \begin{tabular}{c} GCBC \end{tabular} & End-effector & Style variable & 80  & 74\\
   5                 & \begin{tabular}{c} GCBC \end{tabular} & End-effector & None & 78  & 70\\
   6                 & \begin{tabular}{c} GCBC \end{tabular} & End-effector & HCBC$_{(action~only)}$ & 82  & 78\\
   \textbf{7(GHCBC)}  & GCBC & End-effector & HCBC$_{(action + image)}$ & \textbf{96}  & \textbf{92} \\
   \hline
    \end{tabular}
    \caption{Experiments in simulation and real environment.}
    \vspace{-2mm}
    \label{table:ablation2}
\end{table*}

\begin{table}[h]
    \centering
    \scalebox{0.9}{
        \begin{tabular}{cc}
            \hline
            Model parameters & Values \\
            \hline
            Vision encoder & EfficientNet-B0 \\
            Gripper close threshold & 0.6 \\
            Gripper open threshold & 0.4 \\
            HCBC Transformer encoder layers & 4 \\
            Constrained Pose Transformer encoder layers & 4 \\
            Constrained Pose Transformer decoder layers & 7 \\
            Attention heads & 8 \\
            Chunking size & 20 \\
            History length & 20 \\
            \hline
        \end{tabular}
    }
    \caption{Parameters of the model.}
    \label{model-para-table}
\end{table}

\subsection{Result and Discussion}
In order to evaluate the proposed HCBC and GCBC approaches, we designed specific experiments as listed in table \ref{table:ablation2}. SOTA BC method ACT was chosen as a baseline, wherein both input and output joint pose were used, as shown in row 1. 
Row2 replaced joint with end-effector as output pose and row 3 added end-effector pose on input pose, respectively, to study the possible effect of end-effector pose on BC learning. Row 4 and 5 introduced the GCBC model to see how the BC learning efficiency could be improved under various testing conditions. Row 6 and 7 introduced both the GCBC and the HCBC to see overall effect of GHCBC on BC learning performance. 

The parameters of all experiments in table \ref{table:ablation2} were listed in table \ref{model-para-table}. In all experiments, the modules were tested online for 50 times after every 1000 epochs to calculate the success rates until 16000 epochs, and the best success rates during testing were select from the 16 online tests and were listed in table \ref{table:ablation2} which were all on sorting tasks.

\begin{figure*}[h]
    \centering
    \centerline{\subfigure[Sorting task. ]{\includegraphics[width=0.3\linewidth]{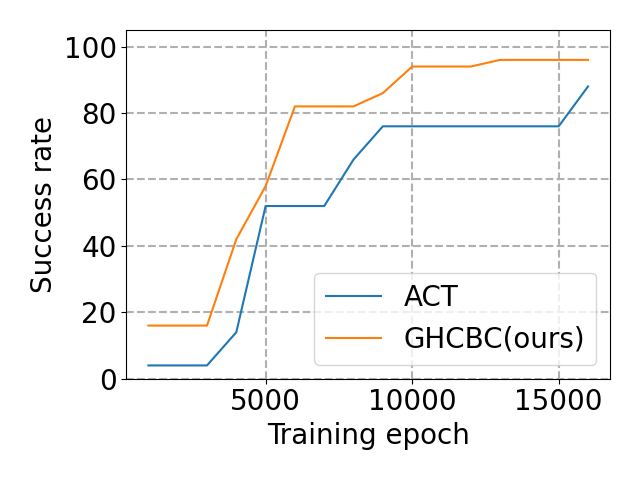}\label{res:sort}}
    \subfigure[Button pushing task. ]{\includegraphics[width=0.3\linewidth]{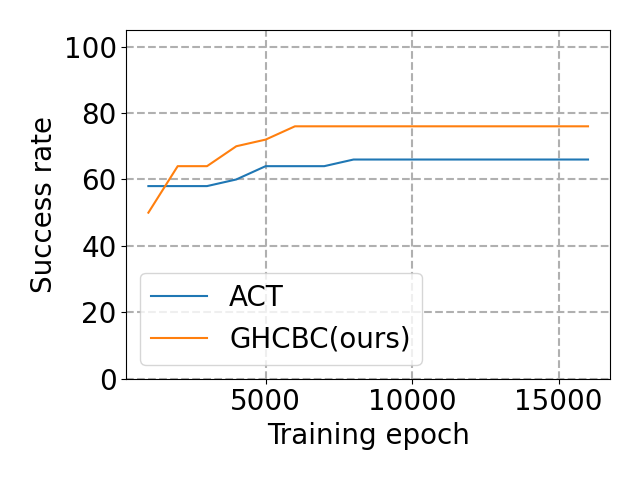}\label{res:push}}
    \subfigure[Basketball shooting tasks. ]{\includegraphics[width=0.3\linewidth]{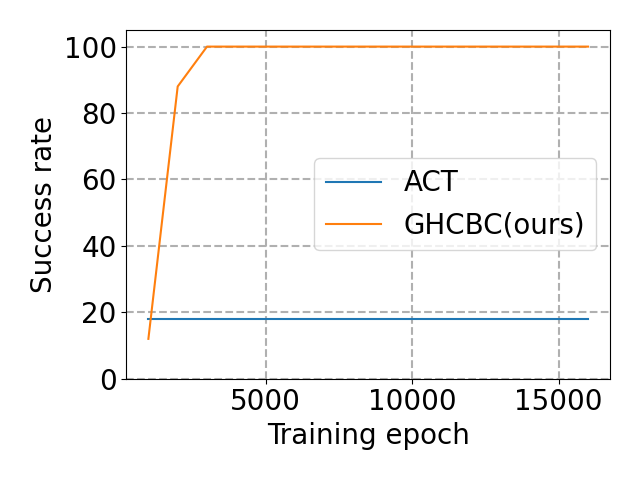}\label{res:basket}}}
    \caption{Success rate of three tasks. }
\end{figure*}

\subsubsection{Effects of pose output. }
The ACT in raw 1 showed baseline success rate of 70\% and 62\% in simulation and real robot experiments, respectively. When the pose output was transformed from joint to end-effector pose, a 4\% success rate improvement was obtained in both simulation and real robot experiments, as shown in raw 2 of table \ref{table:ablation2}. This was partially due to the fact that the wrist camera was located more closely to the end-effector over robot joints in our experiments and the BC module preferred to learn more efficiently when the relationship between wrist camera images and the end-effector pose were used.

\subsubsection{Effects of pose input. }
Effects of pose input.  When joint pose and end-effector pose were used as input simultaneously, it showed a significant decrease of 14\% in simulation and 10\% in robot experiments, respectively, as shown in raw 3 of table \ref{table:ablation2}. It was mainly caused by the attention allocation mechanism intrinsically existed in the BC learning model. When both joint pose and end-effector pose were used as input, BC learning model would pay more attention to robot itself, i.e., joint and end-effector. Meanwhile, it paid less attention to vision information, under the assumption that the attention capability of the BC learning model was fixed. In our experiments, the vision information obtained from the wrist camera was the dominant one guiding how the robot moved, and this made the BC learning capability degrade when vision information was less utilized. 

\subsubsection{Effects of Geometrically Constrained Behavior Cloning (GCBC). }
Inspired by neuroscientists, BC learning efficiency could be improved if high-level perceptual information was used during imitation. In robot learning, this indicated that we might need to dominantly consider relative pose information of joints and end-effectors in order to evaluate the above-mentioned idea. Practically, we introduced geometrical constrain which calculated relative pose information to mathematically design a model to fulfill the evaluation, i.e., GCBC. In fact, the BC learning performance was improved by 33.3\% in simulation and 32.1\% in real robot experiments, respectively, after using the GCBC, as shown in raw 4 of table \ref{table:ablation2}, indicating its superior capability for improving BC learning performance. 

\subsubsection{Effects of Historically Constrained Behavior Cloning (HCBC). }
Based on the GCBC, we further replaced the CVAE Encoder used in \cite{ALOHA:conf/rss/ZhaoKLF23} with the HCBC module to see its effect on BC learning. As shown in raw 6 of table 2, the BC learning performance was improved by 2.5\% in simulation and 5.4\% in real robot experiments respectively, after using the HCBC with only historical pose information, indicating that the copycat issue was effectively eliminated by using the GCBC and HCBC approaches. In fact, the BC learning efficiency was further improved by 17.05\% in simulation and 17.95\% in real robot experiments, respectively, after using the HCBC with both historical pose information and historical vision information. This clearly illustrated that the historical vision information played dominant role in the HCBC method in terms of improving the overall BC learning performance.

In order to evaluate the proposed CBC approach in a clearer manner, we plotted how success rate changed with varying training epochs for both the GHCBC and SOTA BC method.Our experiments showed that the GHCBC always achieved significantly higher success rates on these tasks over the ACT, i.e., 96\% vs. 74\% in the sorting task; 76\% vs. 66\% in the button pushing task, and 100\% vs. 18\% in the basketball shooting task, respectively.

\section{Conclusion}
This paper introduced geometrically and historically constrained behavior cloning (GHCBC), an innovative approach that enhanced traditional behavior cloning (BC) by integrating high-level state information and memory embeddings to address the limitations posed by limited field of view and compound error accumulation.

The proposed GHCBC, which included historical constrained behavioral cloning (HCBC) and geometrically constrained behavioral cloning (GCBC) modules, demonstrated significant improvements in both simulation and real-world experiments.
The HCBC module effectively encoded vision and action history, enabling the model to learn temporal dependencies and enhance its temporal perception capabilities.
Meanwhile, the GCBC module introduced geometrical constraints to focus on current pose information and motion vectors, mitigating the copycat issue and improving BC learning adaptability in dynamic environments.

Comprehensive experimental results on the RLBench benchmark in various tasks such as block sorting, button pushing, and basketball shooting showed that the GHCBC outperformed SOTA BC policy with an average improvement of over 29.74\% in simulation and 39.4\% in real robot experiments, respectively.
Despite its superior performance, GCBC's dependence on complex geometrical information often increased computational complexity when handling high-dimensional geometrical data.
Moreover, in dynamic or non-static environments, geometrical information rapidly changed, rendering decisions based on such inaccurate or ineffective information.
In the future, we will investigate into applying this approach to more complex and dynamic environments.

\bibliography{aaai25}

\appendix
\pagestyle{empty}
\newpage
\section*{AAAI Reproducibility Checklist}

Unless specified otherwise, please answer \answerYes{} to each question if the relevant information is described either in the paper itself or in a technical appendix with an explicit reference from the main paper. If you wish to explain an answer further, please do so in a section titled “Reproducibility Checklist” at the end of the technical appendix.

This paper:
\begin{itemize}
    \item Includes a conceptual outline and/or pseudocode description of AI methods introduced (\answerYes{})
    \item Clearly delineates statements that are opinions, hypothesis, and speculation from objective facts and results (\answerYes{})
    \item Provides well marked pedagogical references for less-familiare readers to gain background necessary to replicate the paper (\answerYes{})
\end{itemize}

Does this paper make theoretical contributions? (\answerYes{})

If yes, please complete the list below.
\begin{itemize}
    \item All assumptions and restrictions are stated clearly and formally. (\answerYes{})
    \item All novel claims are stated formally (e.g., in theorem statements). (\answerYes{})
    \item Proofs of all novel claims are included. (\answerYes{})
    \item Proof sketches or intuitions are given for complex and/or novel results. (\answerYes{})
    \item Appropriate citations to theoretical tools used are given. (\answerYes{})
    \item All theoretical claims are demonstrated empirically to hold. (\answerYes{})
    \item All experimental code used to eliminate or disprove claims is included. (\answerNo[We are organizing the code, and we will make the code publicly available upon publication of the paper with a license that allows free usage for research purposes.])
\end{itemize}

Does this paper rely on one or more datasets? (\answerYes{})

If yes, please complete the list below.
\begin{itemize}
    \item A motivation is given for why the experiments are conducted on the selected datasets (\answerYes{})
    \item All novel datasets introduced in this paper are included in a data appendix. (\answerNo[We introduced a new task (sorting) based on RLBench, and we were organizing the datasets]) 
    \item All novel datasets introduced in this paper will be made publicly available upon publication of the paper with a license that allows free usage for research purposes. (\answerYes{})
    \item All datasets drawn from the existing literature (potentially including authors’ own previously published work) are accompanied by appropriate citations. (\answerYes{})
    \item All datasets drawn from the existing literature (potentially including authors’ own previously published work) are publicly available. (\answerYes{})
    \item All datasets that are not publicly available are described in detail, with explanation why publicly available alternatives are not scientifically satisficing. (\answerNA[])
\end{itemize}
Does this paper include computational experiments? (\answerYes{})

If yes, please complete the list below.
\begin{itemize}
    \item Any code required for pre-processing data is included in the appendix.(\answerNo[We are organizing the code, and we will make the code publicly available upon publication of the paper with a license that allows free usage for research purposes.]).
    \item All source code required for conducting and analyzing the experiments is included in a code appendix. (\answerNo[We are organizing the code, and we will make the code publicly available upon publication of the paper with a license that allows free usage for research purposes.])
    \item All source code required for conducting and analyzing the experiments will be made publicly available upon publication of the paper with a license that allows free usage for research purposes. (\answerYes{})
    \item All source code implementing new methods have comments detailing the implementation, with references to the paper where each step comes from (\answerYes{})
    \item If an algorithm depends on randomness, then the method used for setting seeds is described in a way sufficient to allow replication of results. (\answerYes{})
    \item This paper specifies the computing infrastructure used for running experiments (hardware and software), including GPU/CPU models; amount of memory; operating system; names and versions of relevant software libraries and frameworks. (\answerYes{})
    \item This paper formally describes evaluation metrics used and explains the motivation for choosing these metrics. (\answerYes{})
    \item This paper states the number of algorithm runs used to compute each reported result. (\answerYes{})
    \item Analysis of experiments goes beyond single-dimensional summaries of performance (e.g., average; median) to include measures of variation, confidence, or other distributional information. (\answerYes{})
    \item The significance of any improvement or decrease in performance is judged using appropriate statistical tests (e.g., Wilcoxon signed-rank). (\answerYes{})
    \item This paper lists all final (hyper-)parameters used for each model/algorithm in the paper’s experiments. (\answerYes{})
    \item This paper states the number and range of values tried per (hyper-) parameter during development of the paper, along with the criterion used for selecting the final parameter setting. (\answerYes{})
\end{itemize}

\end{document}